\documentclass[10pt,twocolumn,letterpaper]{article}

\usepackage{cuted}
\usepackage{cvpr}              
\usepackage{multirow}
\definecolor{cvprblue}{rgb}{0.21,0.49,0.74}
\usepackage[pagebackref,breaklinks,colorlinks,allcolors=cvprblue]{hyperref}

\def\paperID{29593} 
\def\confName{CVPR}
\def\confYear{2026}

\title{VGA-Bench: A Unified Benchmark and Multi-Model Framework for Video Aesthetics and Generation Quality Evaluation}

\author{
Longteng Jiang\textsuperscript{1} \quad DanDan Zheng\textsuperscript{1} \quad Qianqian Qiao\textsuperscript{1} \quad Heng Huang\textsuperscript{1} \\ \quad Huaye Wang\textsuperscript{1} \quad Yihang Bo\textsuperscript{2} \quad Bao Peng\textsuperscript{2} \quad Jingdong Chen\textsuperscript{1, \dag} \quad JUN ZHOU\textsuperscript{1} \quad Xin Jin\textsuperscript{3, \dag}
\\
\textsuperscript{1}Ant Group \quad \textsuperscript{2}Beijing Film Academy
\\
\textsuperscript{3}State Key Laboratory of General Artificial Intelligence, BIGAI}

\begin{document}
\maketitle
\let\thefootnote\relax\footnotetext{\textsuperscript{\dag}Corresponding authors.}

\begin{figure*}
  \includegraphics[width=\textwidth]{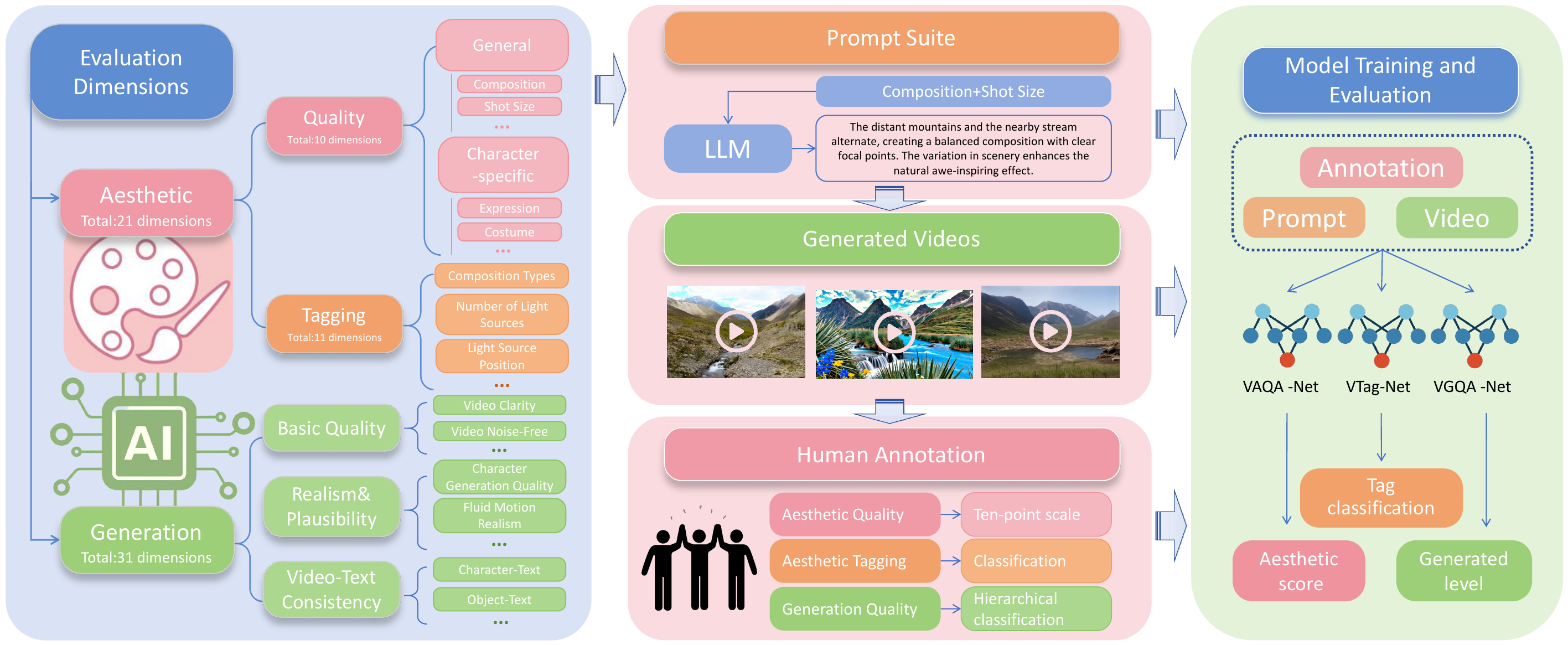}
  \caption{Overview of VGA-Bench. We propose a unified benchmark and multi-model framework for video aesthetic and generation quality assessment, comprising a Prompt Suite with design guidelines, a large-scale generated video dataset, a subset of human-annotated data, and three trained evaluation models—VAQA-Net, VTag-Net, and VGQA-Net—for assessing video aesthetic quality, aesthetic tags, and generation quality, respectively.}
  \label{fig:head}
\end{figure*}

\begin{abstract}

The rapid advancement of AIGC-based video generation has underscored the critical need for comprehensive evaluation frameworks that go beyond traditional generation quality metrics to encompass aesthetic appeal. However, existing benchmarks remain largely focused on technical fidelity, leaving a significant gap in holistic assessment—particularly with respect to perceptual and artistic qualities. To address this limitation, we introduce VGA-Bench, a unified benchmark for joint evaluation of video generation quality and aesthetic quality.

VGA-Bench is built upon a principled three-tier taxonomy: Aesthetic Quality, Aesthetic Tagging, and Generation Quality, each decomposed into multiple fine-grained sub-dimensions to enable systematic assessment. Guided by this taxonomy, we design 1,016 diverse prompts and generate a large-scale dataset of over 60,000 videos using 12 video generation models, ensuring broad coverage across content, style, and artifacts.

To enable scalable and automated evaluation, we annotate a subset of the dataset via human labeling and develop three dedicated multi-task neural assessors: VAQA-Net for aesthetic quality prediction, VTag-Net for automatic aesthetic tagging, and VGQA-Net for generation and basic quality attributes. Extensive experiments demonstrate that our models achieve reliable alignment with human judgments, offering both accuracy and efficiency. We release VGA-Bench as a public benchmark to foster research in AIGC evaluation, with applications in content moderation, model debugging, and generative model optimization.
\end{abstract}    
\section{Introduction}

\label{sec:intro}

In recent years, Artificial Intelligence Generated Content (AIGC) technologies, particularly in the realm of video generation~\cite{blattmann2023stable, ho2022imagen, singer2022make, blattmann2023align, luo2023videofusion, khachatryan2023text2video}, have seen rapid advancements. Leveraging progress in diffusion models~\cite{blattmann2023stable, song2020score, ho2020denoising, zhang2023adding}, transformers~\cite{liu2022video, arnab2021vivit, selva2023video}, and large-scale vision-language pretraining~\cite{chen2023vlp, gan2022vision, wang2023image, dou2022coarse}, current video generation models~\cite{wan2025wan, kong2024hunyuanvideo, tang2025human, liu2024sora, hacohen2024ltx, genmo2024mochi, ma2024latte, yang2024cogvideox, wang2023modelscope, zhang2025show, wang2025lavie, guo2023animatediff, blattmann2023stable} can now produce highly coherent, temporally stable, and visually appealing video sequences from text prompts. These capabilities hold significant potential for applications in digital art, film production, and virtual reality. However, as these generative models become increasingly sophisticated, the need for a comprehensive, reliable, and interpretable evaluation framework becomes more pressing.

Traditional metrics such as FVD~\cite{unterthiner2019fvd}, CLIP Score~\cite{hessel2021clipscore}, or their upgraded versions~\cite{liu2023fetv} primarily focus on technical fidelity—measuring temporal consistency, prompt alignment, or image distortion levels—but often fail to capture higher-level perceptual qualities, especially aesthetic expressiveness that critically influences visual content. Although recent studies have attempted to address this gap, most existing benchmarks still suffer from limited coverage and coarse-grained assessment.

Among them, V-Bench~\cite{huang2024vbench} represents one of the first systematic efforts to evaluate AIGC videos across multiple dimensions, marking an important step towards standardized evaluation. However, it simplifies “video aesthetics” into a single score metric and heavily relies on external scoring models (e.g., MUSIQ~\cite{ke2021musiq}, DINO~\cite{caron2021emerging}), resulting in insufficient granularity, significant bias, and weak controllability.

As shown in Figure \ref{fig:head}, to overcome the aforementioned limitations, this paper introduces VGA-Bench: a unified and fine-grained evaluation benchmark for AIGC-generated videos, aiming to enable joint assessment of generation quality, aesthetic quality, and visual formal elements (tags). Our main contributions are as follows:
\begin{itemize}

\item A detailed and systematic three-dimensional evaluation framework: Building upon V-Bench, we refine the taxonomy by proposing a comprehensive structure encompassing three core dimensions—generation quality, aesthetic quality, and visual formal elements. Each dimension is further decomposed into well-defined sub-attributes (e.g., composition, color harmony, lighting usage, motion aesthetics), enabling fine-grained, interpretable, and holistic evaluation.

\item A diverse prompt suite and large-scale test dataset: We design 1,016 diverse prompts based on the evaluation framework, covering various scenes, actions, styles, and challenging scenarios. Using 12 state-of-the-art video generation models, we generate a total of 60,000 videos, constructing the largest integrated testing platform to date and supporting fair cross-model comparisons.

\item Three dedicated multi-task automated evaluators: Based on professional human annotations, we train three specialized neural assessors—VGQA-Net (for generation quality prediction), VAQA-Net (for aesthetic quality assessment), and VTag-Net (for automatic aesthetic tagging)—eliminating reliance on external scoring models and enabling end-to-end, consistent, and scalable automated evaluation.

\item Comprehensive empirical analysis of mainstream models with full open-source commitment: We conduct a systematic evaluation of 12 cutting-edge models using VGA-Bench, revealing their strengths and weaknesses across different dimensions. Upon publication, we will fully release: (1) the complete benchmark suite (including taxonomy, prompt templates, and annotation data); (2) public API interfaces for all evaluation models; (3) the entire generated video dataset—ensuring reproducibility and broad accessibility for the research community.
\end{itemize}

We believe that VGA-Bench serves not only as a rigorous evaluation platform but also as a key infrastructure for advancing the next generation of video generation systems with enhanced aesthetic intelligence and artistic controllability.
\section{Related Work}
\label{sec:related}

\begin{table*}
    \centering
    \begin{tabular}{l c c c c }
    \hline
     & Total Dimensions & Aesthetic Dimensions & Evaluated Models & Prompts \\
    \hline
    VBench~\cite{huang2024vbench}   & 16 & 1  & 4 & $\sim$1600\\
    VBench2.0~\cite{zheng2025vbench}   & 18 & 2  & 4 & $\sim$1600\\
    T2V-CompBench~\cite{sun2025t2v}   & 7 & 0  & 23 & 1400\\
    ChronoMagic-Bench~\cite{yuan2024chronomagic}   & 4 & 0  & 13 & 1649\\
    StoryEval~\cite{wang2025your}   & 8 & 0  & 11 & 423\\
    VGA-Bench(ours)   & 52 & 21  & 12 & 1016\\
    \hline
    \end{tabular}
    \caption{Comparison of existing evaluation methods for text-to-video generative models}
    \label{tab:biao1}
\end{table*}

\subsection{Video Generative Models}
In recent years, driven by the rapid advancement of deep generative models, text-to-video (T2V) generation technology~\cite{blattmann2023stable, ho2022imagen, singer2022make, blattmann2023align, luo2023videofusion, khachatryan2023text2video} has achieved significant breakthroughs. Generative architectures represented by diffusion models~\cite{song2020score, ho2020denoising, zhang2023adding, blattmann2023stable, li2022dit} are now capable of producing high-resolution, temporally coherent, and creatively rich dynamic content from natural language descriptions. These technologies not only show broad application prospects in film production, advertising design, game development, and virtual reality, but are also increasingly integrated into social media creation and personalized content generation pipelines, becoming a core component of the AIGC ecosystem.

Previously, mainstream T2V model architectures were dominated by U-Net-based designs~\cite{wang2023modelscope, guo2023animatediff, blattmann2023stable}, but their limitations---such as difficulties in modeling long-range dependencies and poor scalability---soon became apparent. With the emergence of Sora~\cite{liu2024sora}, pure Transformer-based architectures exemplified by DiT~\cite{ma2024latte, yang2024cogvideox, li2022dit} have rapidly gained prominence due to their unparalleled global modeling capability and excellent scalability, and are now becoming the dominant paradigm and future direction for high-end, large-scale video generation models.

However, as generation capabilities improve, user expectations have evolved beyond basic technical correctness (e.g., absence of artifacts, plausible motion) to increasingly emphasize artistic expressiveness and aesthetic quality---such as whether the composition is visually pleasing, lighting is skillfully employed, color harmony is well-balanced, or character expressions are graceful. At the same time, the fidelity with which generated content reflects key visual elements described in the prompt (e.g., ``a cyberpunk cityscape at night'' or ``a slow-motion dance under soft backlighting'')---i.e., consistency in visual formal elements---has become a crucial metric for assessing model controllability and semantic understanding.

In this paper, we evaluate a series of text-to-video models released over the past three years, including both officially open-sourced and commercial models. This comprehensive evaluation ensures diversity in T2V approaches and provides insightful analysis into their capabilities.

\subsection{Evaluation of Video Generative Models}
Despite continuous performance improvements in video generation models, the scientific and fair evaluation of their comprehensive capabilities remains a key challenge in research. Early assessment methods primarily relied on human ratings or simple technical metrics~\cite{unterthiner2019fvd, hessel2021clipscore, liu2023fetv}, which are insufficient for quantifying complex human perceptual experiences. In recent years, with the development of the AIGC ecosystem, a series of specialized evaluation benchmarks for text-to-video generation have been proposed, driving the evolution of assessment frameworks from single metrics toward multi-dimensional and automated paradigms.

Among them, V-Bench~\cite{huang2024vbench} is the first comprehensive benchmark for video generation, decomposing evaluation into multiple sub-tasks---including visual quality, prompt alignment, and motion plausibility---and incorporating human annotations for holistic scoring. Its successor, V-Bench2~\cite{zheng2025vbench}, extends the original framework by introducing additional dimensions such as generated duration, frame rate, and style diversity, while also incorporating more generative models and test samples to enhance evaluation breadth and representativeness. Subsequent benchmarks such as ChronoMagic-Bench~\cite{yuan2024chronomagic}, T2V-CompBench~\cite{sun2025t2v}, and StoryEval~\cite{wang2025your} have also evaluated model performance from multiple perspectives, covering aspects like temporal coherence, compositional fidelity, and narrative consistency. Table \ref{tab:biao1} presents a comparative overview of key data characteristics between our work and these existing benchmarks.

However, existing benchmarks still suffer from several limitations:

\begin{itemize}
    \item Over-simplified aesthetic evaluation: Most benchmarks treat ``aesthetic quality'' as a single holistic metric, lacking fine-grained modeling of specific aesthetic elements such as composition, color harmony, lighting, and visual rhythm.

    \item Reliance on external models: Many metrics depend on pre-trained image or video understanding models for indirect inference, which may introduce bias and fail to align with genuine human perception.
\end{itemize}

To address these shortcomings, this paper proposes VGA-Bench, which introduces systematic improvements at three levels: evaluation dimensions, data construction, and model design. We not only refine the sub-dimensions of aesthetic quality but also develop dedicated multi-task evaluation models specifically designed for video aesthetics and generation quality, enabling more comprehensive and accurate assessment of generated videos.
\section{VGA-Bench Suite}
\label{sec:Suite}

\subsection{Evaluation Dimension Suite}
\subsubsection{Aesthetic Quality}
Video aesthetics refers to the perceptual appeal and artistic expressiveness conveyed through visual formal elements—such as composition, color, lighting, and motion—in artificially generated dynamic content. Our aesthetic quality dimensions are adapted from the VADB dataset~\cite{qiao2025vadb}, and specifically include the following ten dimensions: Overall Score, Composition, Shot Size, Lighting, Visual Tone, Color, Depth of Field, Expression, Costume, and Makeup. The definitions of these dimensions in real-world videos are thoroughly described in the original dataset paper and thus will not be repeated here. Instead, we focus on their manifestation and interpretability within the context of generated videos.

    \textbf{Composition (Com):} Generated videos often suffer from ``floating composition'' or ``visual center offset'' due to a lack of spatial logic, yet they can achieve surreal arrangements that are difficult to realize in real-world filming.
    
    \textbf{Shot Size (SS):} In real videos, shot selection is constrained by physical camera setups, whereas generated videos allow free perspective switching—but sometimes lack narrative coherence in ``cinematic language.''
    
    \textbf{Lighting (Lig):} Real-world lighting appears natural and physically plausible, while generated videos may exhibit ``uniform illumination'' or ``non-physical light sources,'' leading to stylized yet distorted appearances.
    
    \textbf{Visual Tone (VT):} Generated videos demonstrate more consistent tone control, but tend toward ``template-like emotional expression'' and lack the subtle transitions present in real lighting dynamics.
    
    \textbf{Color (Col):} Colors in real videos are rich and influenced by environmental conditions, whereas generated videos often adopt an ``idealized'' palette, frequently exhibiting stylistic biases such as ``over-saturation'' or ``low contrast.''
    
    \textbf{Depth of Field (DoF):} In real footage, depth of field dynamically changes with focus; in contrast, generated videos often feature ``static blur'' effects, lacking the dynamic perception of spatial layers.
    
    \textbf{Expression (Exp):} Real performances contain micro-expressions and emotional fluctuations, while expressions in generated videos are often ``mechanical'' or ``stiff,'' failing to capture complex psychological states.
    
    \textbf{Costume (Cos):} Costumes in real videos are grounded in cultural and historical context, whereas generated videos frequently produce ``style mismatches'' or ``inappropriate attire'' due to inconsistent semantic reasoning.
    
    \textbf{Makeup (Mak):} Real makeup emphasizes detail fidelity and skin-tone harmony, while generated videos often exhibit ``texture discontinuities'' or ``proportional distortions'' in virtual makeup rendering.

We define these aesthetic quality dimensions to guide generative models toward the high-level aesthetic standards observed in real-world videos, enabling systematic evaluation of their alignment with human perception in aspects such as composition, lighting, and color. Through fine-grained aesthetic assessment, we aim to examine the model's understanding and reconstruction ability regarding advanced visual aesthetics, thereby promoting AIGC systems to more deeply grasp and generate high-quality content that aligns with human aesthetic preferences.

\subsubsection{Aesthetic Tagging}
Aesthetic video tags are structured annotations of identifiable and quantifiable visual aesthetic features in a video, used to describe artistic expression elements such as composition style, lighting application, and color properties. Similarly, our aesthetic video tags are adapted from the VADB dataset~\cite{qiao2025vadb} and supplemented by established photographic theory~\cite{matbouly2022quantifying, deren1960cinematography, brown2016cinematography}. We select the following 11 aesthetic tags: Composition Types, Number of Light Sources, Light Source Position, Light Quality, Light Color, Shot Type, Depth of Field, Saturation, Brightness, Color Temperature, and Contrast. Definitions for each dimension are provided below:

    \textbf{Composition Types (CT):} Refers to the spatial arrangement of the main subject and visual elements within the frame, influencing visual balance and narrative guidance. Includes: Rule of Thirds Composition, Symmetrical Composition, Asymmetrical Composition, Centered Composition, Framing Composition, Leading Lines Composition.
    
    \textbf{Number of Light Sources (NoLS):} The number of primary illumination sources in the scene, affecting depth perception, atmosphere, and spatial layering. Includes: Single Light Source, Dual Light Sources, Multiple Light Sources.
    
    \textbf{Light Source Position (LSP):} The direction of the light relative to the subject, shaping contours, volume, and emotional tone. Includes: Back Light, Front-Side Light, Side Light, Bottom Light, Top Light, Front Light, Back-Side Light.
    
    \textbf{Light Quality (LQ):} The hardness or softness of light—soft light is diffused and even, hard light is sharp and directional—directly influencing mood, texture rendering, and visual texture expression. Includes: Hard Light, Soft Light, Diffused Light.
    
    \textbf{Light Color (LC):} The chromatic property of the light source, used to convey emotion, indicate time of day, or create stylized atmospheres. Includes: White (Neutral) Light, Warm Light, Cool Light, Colored Light.
    
    \textbf{Shot Type (ST):} The distance relationship between the camera and the subject, determining information density and psychological engagement with the viewer. Includes: Wide Shot, Full Shot, Medium Shot, Close-Up, Extreme Close-Up.
    
    \textbf{Depth of Field (DoF):} The range of spatial area that appears in focus; shallow depth of field emphasizes the subject, while deep depth of field reveals environmental context—serving as a key tool for directing visual attention. Includes: Shallow DOF, Deep DOF.
    
    \textbf{Saturation (Sat):} The intensity or purity of colors—high saturation appears vivid and striking, low saturation conveys subtlety and restraint—impacting visual impact and emotional expression. Includes: High, Medium, Low.
    
    \textbf{Brightness (Bri):} The overall luminance level of the image, affecting readability, mood, and perceived spatial depth. Includes: Bright, Medium, Dark.
    
    \textbf{Color Temperature (Col):} The warmth or coolness of the lighting, a critical factor in establishing emotional tone and temporal cues (e.g., dawn vs. dusk). Includes: Cool, Medium, Warm.
    
    \textbf{Contrast (Con):} The difference between the brightest and darkest regions in the image—high contrast enhances dramatic tension, while low contrast creates a soft, harmonious feel. Includes: High, Medium, Low.
    
We define these aesthetic video tags to construct an interpretable and reproducible visual aesthetic language system, enabling evaluation to move beyond subjective judgments such as ``whether it looks good,'' toward concrete analysis of where the visual appeal lies and why it is aesthetically effective. Through standardized tag annotation, we can effectively measure a model’s understanding of photographic aesthetic principles, and provide training signals and optimization objectives for future generation of high-quality videos that better align with human aesthetic preferences.

\subsubsection{Generation Quality}
Our generation quality assessment further refines the framework of V-Bench~\cite{huang2024vbench} by categorizing it into three broad categories comprising a total of 31 sub-dimensions. Video-Text Consistency measures the semantic alignment between the generated content and the input prompt; Reality \& Plausibility evaluates the credibility of scenes, actions, and physical dynamics with respect to real-world laws; Basic Quality focuses on the intrinsic visual clarity and technical stability of the video itself. 

The Video-Text Consistency dimension includes: Character-Text Consistency (1), Action-Text Consistency (2), Scene-Text Consistency (3), Object Position-Text Consistency (4), Object Attribute-Text Consistency (5), Object-Text Consistency (6), Video Content-Text Consistency (7), Video Speed-Text Consistency (8), Video Style-Text Consistency (9), Camera Movement-Text Consistency (10), Unrealistic Description Imaginative Presentation (11).

The Realism \& Plausibility dimension includes: Rigid Body Collision Realism (12), Action Realism (13), Scene Realism (14), Weather Representation Realism (15), Time Period Representation Realism (16), Gaseous Motion Realism (17), Fluid Motion Realism (18), Gradual Change Motion Realism (19), Object Motion Trajectory Realism (20), Object Realism (21), Character Generation Quality (22), Textual Attribute Representation Realism (23), Video Lighting and Shadow Realism (24), Moving Scene Reasonableness (25), Overall Realism (26).

The Basic Quality dimension includes: Abnormal Lighting Detection (27), Video Noise-Free (28), Video Clarity (29), Static Content Non-distortion (30), Static Content Stability (31).

The definitions of all dimensions are summarized in Appendix.

We define these three categories of generation quality and their sub-dimensions to systematically evaluate AIGC videos in terms of semantic understanding, physical commonsense, and visual fidelity. This ensures that models not only ``understand'' the input prompts but also generate content that is logically coherent and visually natural. Through fine-grained decomposition, our framework provides clear optimization directions for model improvement, and promotes the evolution of generative systems toward greater realism, controllability, and practical usability.

\subsection{Prompt Suite}
\subsubsection{Prompt Design}
Prompt design is a critical component in text-to-video evaluation. A clear and precise prompt can effectively reduce stochastic interference during generation, enabling the model to focus on user intent and thus more faithfully reflect its semantic understanding and content generation capabilities. To this end, our core design principle is: the targeted aesthetic or generation quality dimension must be explicitly specified in the prompt, ensuring that the model can perceive and respond to the intended attribute. For example, a video should only be used for composition assessment if the prompt explicitly includes descriptions such as ``composed using the rule of thirds''; otherwise, the corresponding dimension should not be included in the evaluation.

Building upon this, we further emphasize prompt diversity: prompts should vary in length, cover both single-dimension and multi-dimensional scenarios, and span a wide range of themes and scenes to enhance the representativeness and robustness of the test set.

Based on these principles, we construct a systematic Prompt Suite containing 1,016 carefully designed prompts, distributed as follows: 200 for aesthetic quality dimensions, 220 for aesthetic tag dimensions, and 596 for generation quality dimensions. Each evaluation dimension is covered by at least 50 prompts, ensuring statistical validity. Furthermore, to accommodate different testing requirements, we provide two lightweight subsets: one with 508 prompts and another with 127 prompts. All versions maintain balanced dimension distribution and diverse prompt lengths, and support combinations of 1 to 5 dimensions per prompt, facilitating flexible fine-grained analysis and efficient lightweight evaluation.

\begin{figure}[t]
  \includegraphics[width=\linewidth]{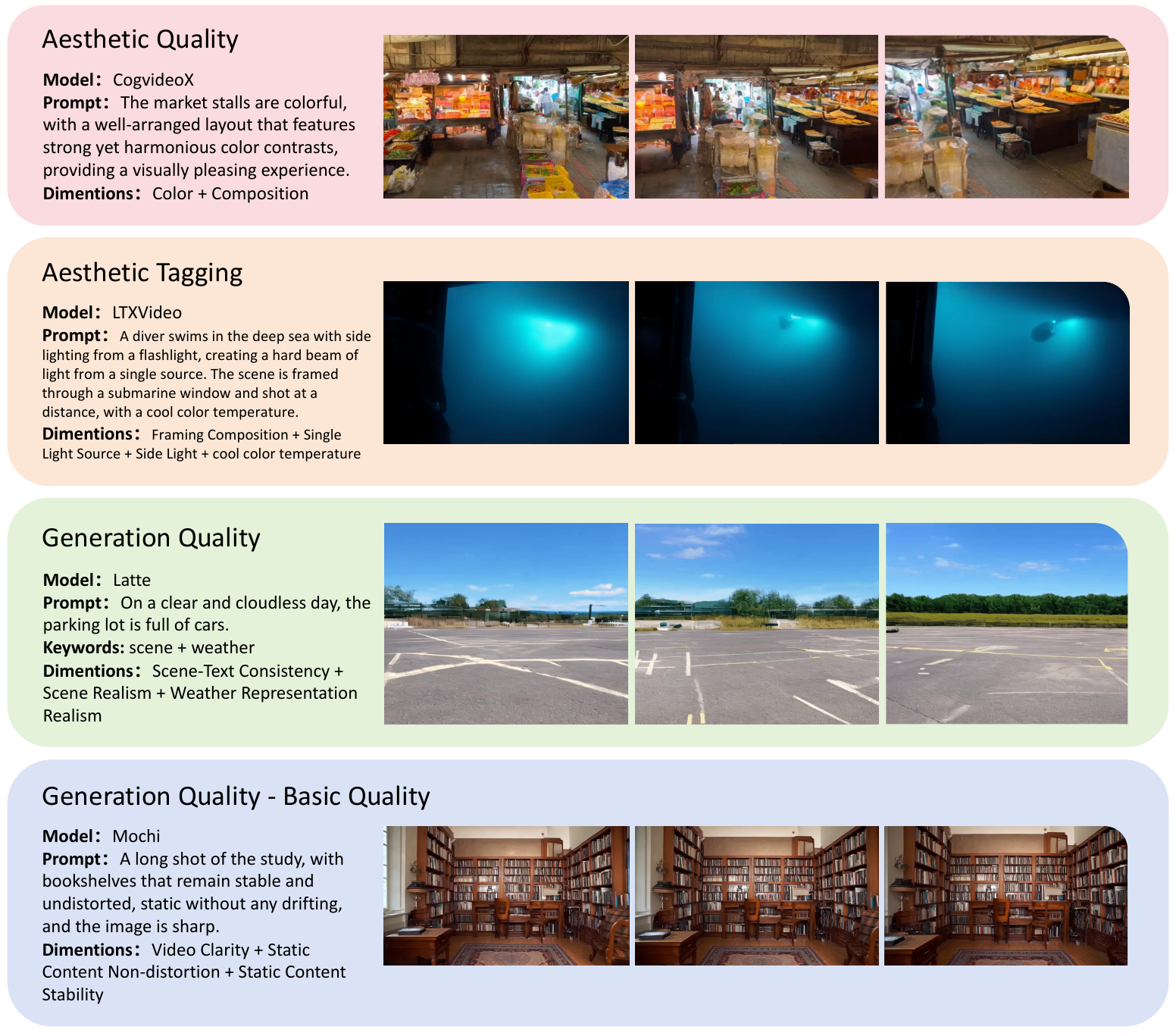}
  \caption{Prompts for the three core dimensions, their corresponding sub-dimensions, and example generated videos.}
  \label{fig:sample-A}
\end{figure}

\subsubsection{Use of LLMs}
For the aesthetic quality dimensions, we select high-scoring real video comments from the VADB dataset~\cite{qiao2025vadb} in the corresponding dimensions and extract descriptive sentences that emphasize specific aesthetic attributes (e.g., ``balanced composition'', ``soft lighting''). These human aesthetic feedbacks are used as input to guide the LLM in generating prompts with similar expressive styles and semantic focus.

For the aesthetic tag dimensions, we directly feed the categorical labels of each sub-dimension (e.g., Back Light, Shallow DOF, High Saturation) into the LLM, instructing it to generate natural language descriptions that explicitly include the given keyword while maintaining semantic coherence.

For the generation quality dimensions, we first summarize each sub-dimension into one or more representative keywords—for example, ``Object'' for both Object-Text Consistency and Object Realism, and ``Gaseous Motion'' for Gaseous Motion Realism—ensuring that each keyword covers one or two core attributes. Subsequently, these keywords are used to prompt the LLM to generate text inputs that precisely elicit the target characteristics. Notably, for the Basic Quality sub-dimensions, we set the keywords as single adjectives and directly incorporate them into the prompt as modifiers—for instance, ``Video Clarity'' is realized in the prompt as a directive such as ``generate a clear video''.

Concrete examples are illustrated in Figure \ref{fig:sample-A}.

\begin{figure}[t]
  \includegraphics[width=\linewidth]{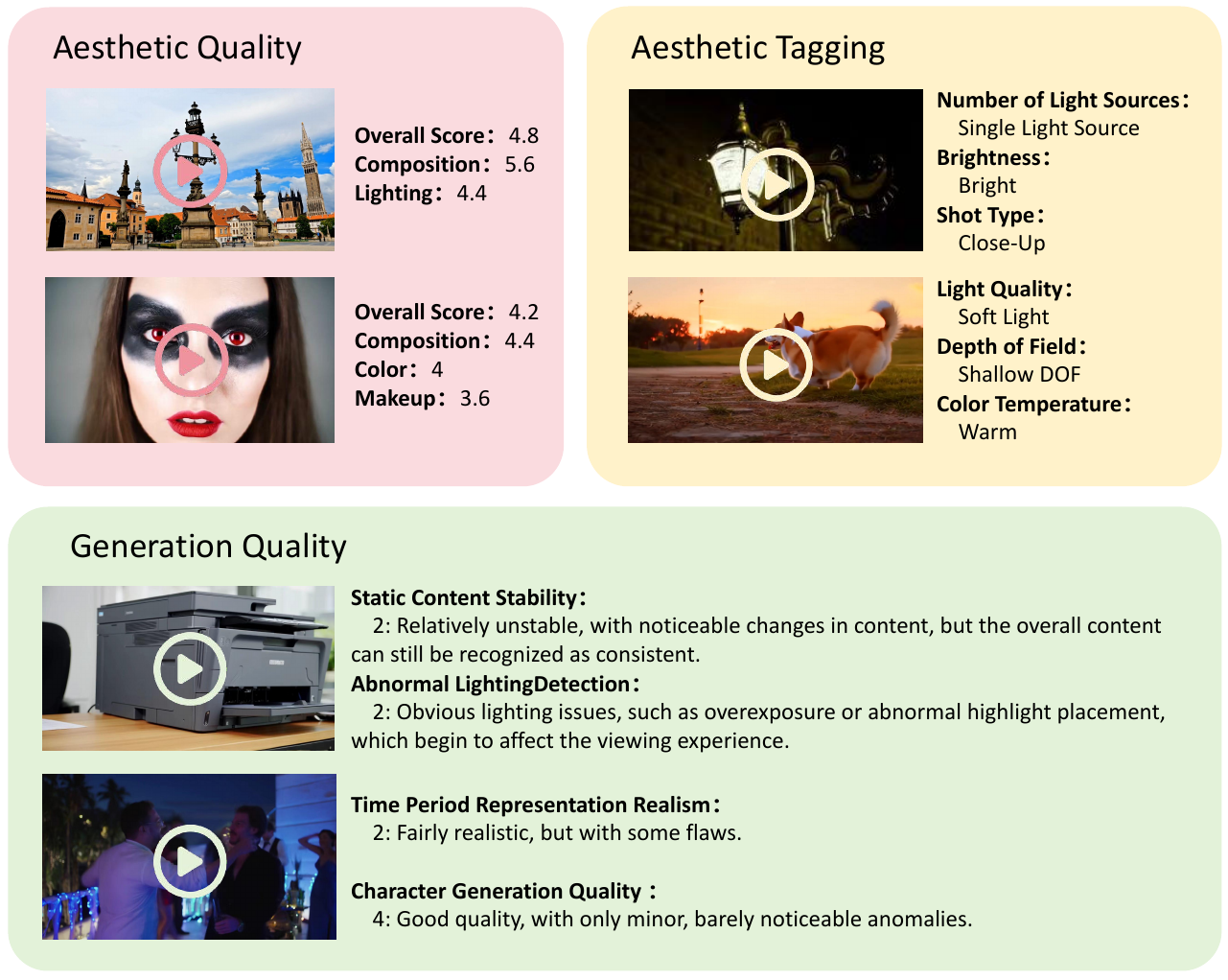}
  \caption{Examples of human annotations for the three core dimensions.}
  \label{fig:sample-B}
\end{figure}

\subsection{Human Annotation}
To train the multi-task evaluation models of this benchmark—VGQA-Net, VAQA-Net, and VTag-Net—we adopt an ``expert-led + crowd-assisted'' annotation paradigm. First, domain experts from the film and video industry perform exemplary annotations on a subset of samples based on predefined dimension definitions and rating guidelines. Subsequently, a crowdsourced team completes the labeling of the remaining data by following these exemplars. Finally, experts conduct batch-wise sampling audits; if annotation errors are identified, the entire batch is rejected and re-labeled to ensure consistent quality. All annotations strictly follow the prompt design principles: raters score only those dimensions explicitly mentioned in the prompt, avoiding subjective inference on unmentioned attributes.

For the aesthetic quality and aesthetic tag dimensions, we adopt the standardized scoring guidelines from the VADB dataset~\cite{qiao2025vadb}, with rating boundaries defined through both textual descriptions and example videos. Each aesthetic sub-dimension is scored on a 0–10 scale, and the final score is computed as the average of three independent annotators. For aesthetic tags, treated as a multi-label classification task, each sample is independently labeled by three annotators, and labels are retained only if at least two agree (``majority voting'').

For each evaluation dimension under generation quality, we design specific assessment questions paired with structured response options. These options represent distinct levels of quality—functioning effectively as ordinal scores—tailored to the semantic meaning of the respective dimension.

Example (Object-Text Alignment): This dimension includes four response options:

\begin{itemize}
    \item -1 (Invalid Question): A universal option present in most dimensions. Annotators select this to discard a sample when: The prompt for a ``consistency'' dimension lacks a specified target (e.g., object, scene). The prompt for a ``realism'' dimension intentionally describes an unrealistic scenario.
    \item 1 (Completely Inconsistent): The object exhibits no alignment with the text description.
    \item 2 (Partially Consistent): The object exhibits characteristics of the described target.
    \item 3 (Fully Consistent): The object perfectly matches the text description.
\end{itemize}
    
Likewise, the results for generation quality are determined using the ``majority voting'' principle.
\section{Experiments and Results}
\label{sec:Experiments}

\subsection{VGA Evaluation Network}
\begin{figure}[t]
  \includegraphics[width=\linewidth]{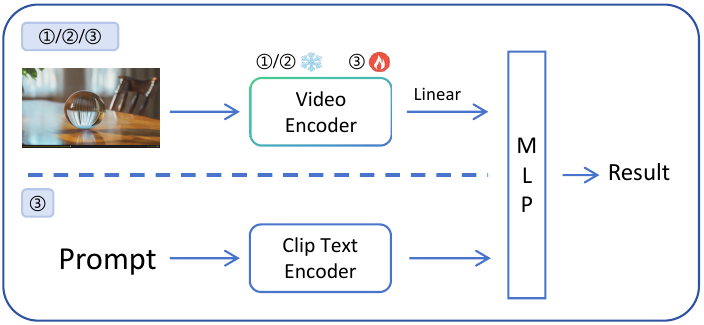}
  \caption{Architecture of (1)VAQA-Net, (2)VTag-Net, and (3)VGQA-Net. The video encoders in VAQA-Net and VTag-Net are those trained in the first stage of the VADB dataset~\cite{qiao2025vadb} using a dual-text encoder with dynamic fusion module for language comments and aesthetic tags; these encoders are frozen during the training phase in this work. Compared to the former two models, VGQA-Net includes an additional CLIP~\cite{radford2021learning} branch before the input MLP.}
  \label{fig:model}
\end{figure}

The network architectures of VAQA-Net, VTag-Net, and VGQA-Net are illustrated in Figure ~\ref{fig:model}.

Despite significant progress in visual fidelity of generated videos, they still lag far behind human-created content in core aspects of aesthetic intelligence—such as intentional artistry, emotional authenticity, and cultural context embedding. Real-world videos, especially professionally produced films and documentaries, exhibit deliberate artistic decisions in composition, lighting, and narrative pacing, reflecting a deep understanding of human perception and cultural norms. These qualities make real-world video data an indispensable resource for training models to recognize and reason about ``meaningful beauty,'' going beyond merely capturing superficial visual patterns.

Therefore, we initialize VAQA-Net and VTag-Net with the video encoder pre-trained in the first stage of the VADB dataset~\cite{qiao2025vadb}, along with its associated training data and parameters from real video scoring and tagging tasks, enabling the models to inherit aesthetic understanding acquired from professional cinematography. In the second stage, we fine-tune the models on an extended dataset that includes 1,300 generated videos from 12 mainstream generative models, each paired with high-quality human annotations. Evaluation is conducted on a separate set of 400 generated videos: for the aesthetic tag task, standard accuracy (Acc) is used as the metric; for aesthetic scoring, the 0–10 scale is discretized into five levels, and five-class accuracy is computed. Results are presented in Table~\ref{tab:biao3} and Table~\ref{tab:biao4}.

\begin{table}
	\centering
    \small
	\caption{5-Class Accuracy of VAQA-Net}
	\label{tab:biao3}
		\begin{tabular}{cc|cc} 
			\hline 
			\textbf{Dim.}  & \textbf{Acc.↑ \%} &\textbf{Dim.}  & \textbf{Acc.↑ \%}\\ \hline
			Overall & 76.9  & Com  &  73.6\\ \hline
			SS & 72.4 &  Lig &  69.5\\ \hline
			VT & 67.9 &  Col &  71.1\\ \hline
			DoF & 69.5 & Exp  &  74.8\\ \hline
			Cos & 77.4 & Mak &  71.8\\ \hline
		
	\end{tabular}
\end{table}

\begin{table}
	\centering
    \small
	\caption{Accuracy of VTag-Net (Top-2 Predicted Tags Match Ground Truth)}
	\label{tab:biao4}
		\begin{tabular}{cc|cc} 
			\hline 
			\textbf{Dim.}  & \textbf{Acc.↑ \%} &\textbf{Dim.}  & \textbf{Acc.↑ \%}\\ \hline
			CT & 45  & NoLS  &  77\\ \hline
			LSP & 66 &  LQ &  74\\ \hline
			LC & 80 & ST &  57\\ \hline
			DoF & 82 & Sat  &  89\\ \hline
			Bri & 82 & Col &  71\\ \hline
            Con & 90 &  &  \\ \hline
		
	\end{tabular}
\end{table}

In contrast, our VGQA-Net is fully focused on generated videos. To comprehensively evaluate its cross-model generalization capability, we select two representative generative models from each year over three years, resulting in six models in total: HunyuanVideo~\cite{kong2024hunyuanvideo}, LTXVideo~\cite{hacohen2024ltx}, Mochi~\cite{genmo2024mochi}, Latte-1~\cite{ma2024latte}, CogVideoX~\cite{yang2024cogvideox}, and Show-1~\cite{zhang2025show}, covering 12,000 generated videos. The model is trained on videos produced by three of these models and tested on videos from the remaining three, ensuring no overlap between training and test sets in terms of model provenance. Accuracy (Acc) is used as the evaluation metric, and results are presented in Table~\ref{tab:biao5}.

\begin{table}
	\centering
	\caption{Accuracy of VGQA-Net}
    \small
	\label{tab:biao5}
		\begin{tabular}{cc|cc|cc} 
			\hline 
			\textbf{Num.}  & \textbf{Acc.↑ \%} &\textbf{Num.}  & \textbf{Acc.↑ \%}
            &\textbf{Num.}  & \textbf{Acc.↑ \%}\\ \hline
			1 & 86.0  & 12  &  82.3 & 23  &  83.0\\ \hline
			2 & 78.4 &  13 &  91.0 & 24  &  82.1\\ \hline
			3 & 78.9 & 14 &  92.6 & 25  &  80.7\\ \hline
			4 & 76.3 & 15  &  81.6 & 26  &  85.6\\ \hline
			5 & 72.4 & 16 &  87.8 & 27  &  89.9\\ \hline
            6 & 74.8 & 17 &  84.4 & 28  &  81.2\\ \hline
            7 & 74.3  & 18  &  71.7 & 29  &  85.6\\ \hline
            8 & 75.8  & 19  &  70.1 & 30  &  76.5\\ \hline
            9 & 76.7  & 20  &  71.2 & 31  &  81.8\\ \hline
            10 & 81.4  & 21  &  83.5 &   &  \\ \hline
            11 & 67.3  & 22  &  80.6 &   &  \\ \hline
		
	\end{tabular}
\end{table}

\begin{figure}[t]
  \includegraphics[width=\linewidth]{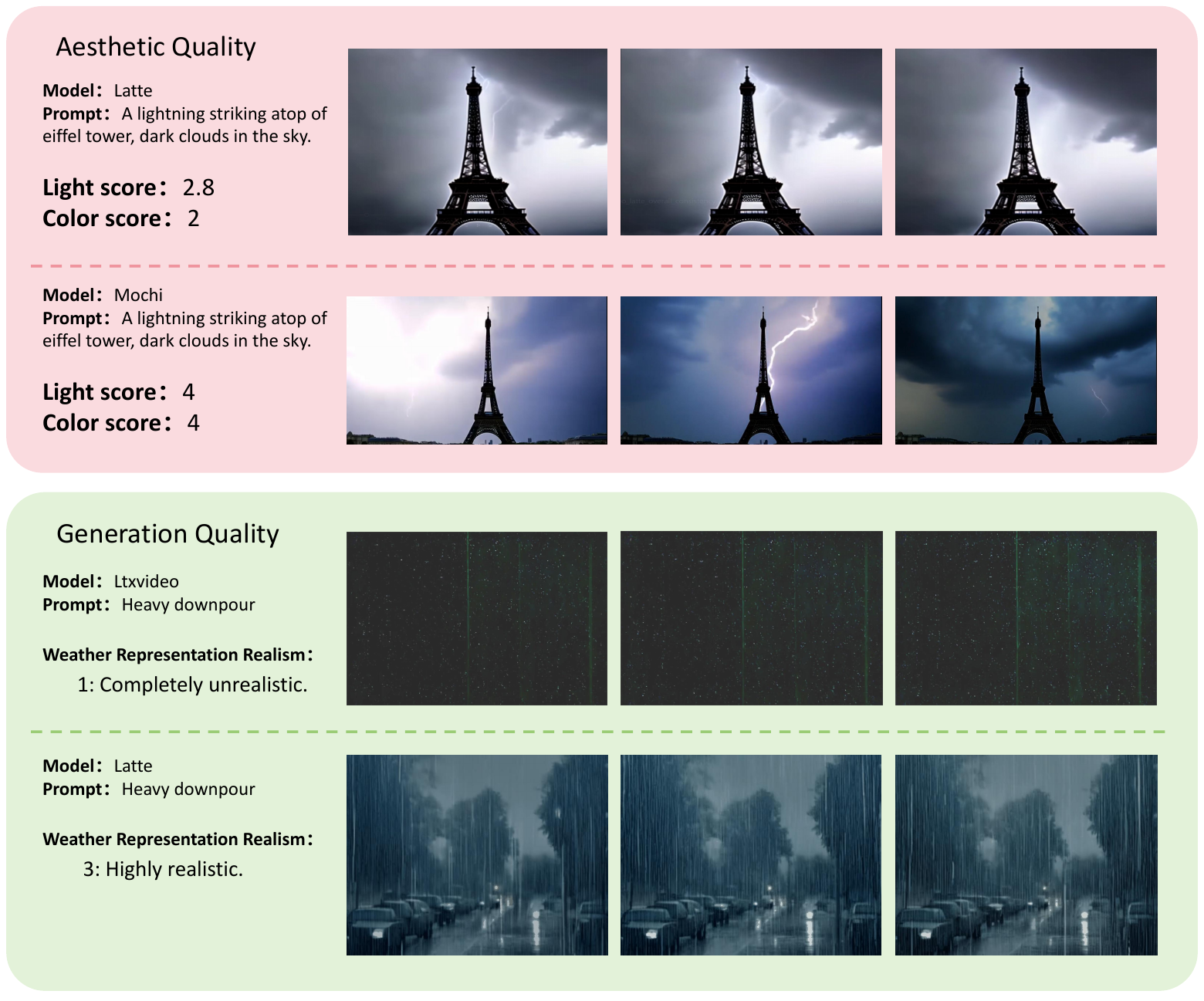}
  \caption{Comparison of generated videos from different models using the same dimension-aligned prompt. Aesthetic scores and generation quality levels are derived from human annotations.}
  \label{fig:sample-C}
\end{figure}

\subsection{VGA-Bench Evaluation Results}
We evaluate all generative models on the trained VAQA-Net, VTag-Net, and VGQA-Net to assess their performance across various aesthetic and quality dimensions. To ensure a fair and unbiased ranking, all generated videos used for evaluation are held out from the model training process, preventing data leakage from influencing the results. The evaluated models are ordered by release date in ascending order, including: Stable Video Diffusion(SVD)~\cite{blattmann2023stable}, AnimateDiff-v2~\cite{guo2023animatediff}, LaVie~\cite{wang2025lavie}, Show-1~\cite{zhang2025show}, ModelScope~\cite{wang2023modelscope}, CogVideoX~\cite{yang2024cogvideox}, Latte-1~\cite{ma2024latte}, Mochi~\cite{genmo2024mochi}, LTXVideo~\cite{hacohen2024ltx}, HunyuanVideo~\cite{kong2024hunyuanvideo}, Wan2.1~\cite{wan2025wan}, and Sora2~\cite{liu2024sora}.

For the aesthetic quality dimension, models are ranked by average score, with higher scores indicating better performance. For the aesthetic tag dimension, classification accuracy is used as the metric, and models with higher mean accuracy rank higher. For the generation quality dimension, models are ranked by average level rating, where a higher average level indicates better generation quality and thus a better (higher) rank.

The results are obtained by normalizing and averaging across all sub-dimensions within the three main dimensions, as shown in Table~\ref{tab:biao6}. For more detailed results of each model across all sub-dimensions, please refer to the Appendix.

\begin{table}

    \centering

    \small 

    \begin{tabular}{l c c c}

    \hline

     Model & Aes. Score & Tag Cla. & Gen. Level \\

    \hline

    SVD~\cite{blattmann2023stable}             & 0.20 & 0.32 & \textbf{0.55} \\

    AnimateDiff~\cite{guo2023animatediff}      & 0.36 & 0.30 & 0.49 \\

    LaVie~\cite{wang2025lavie}                 & 0.34 & 0.31 & 0.47 \\

    Show-1~\cite{zhang2025show}                & 0.29 & 0.32 & 0.28 \\

    ModelScope~\cite{wang2023modelscope}       & 0.31 & 0.30 & 0.49 \\

    CogVideoX~\cite{yang2024cogvideox}         & 0.41 & \textbf{0.39} & \textbf{0.55} \\

    Latte-1~\cite{ma2024latte}                 & 0.35 & 0.31 & 0.45 \\

    Mochi~\cite{genmo2024mochi}                & 0.21 & 0.34 & \underline{0.54} \\

    LTXVideo~\cite{hacohen2024ltx}             & 0.22 & 0.34 & 0.47 \\

    Hunyuan~\cite{kong2024hunyuanvideo}        & 0.45 & 0.36 & \textbf{0.55} \\

    Wan2.1~\cite{wan2025wan}                   & \underline{0.46} & \underline{0.38} & 0.53 \\

    Sora2~\cite{liu2024sora}                   & \textbf{0.50} & 0.18 & \underline{0.54} \\

    \hline

    \end{tabular}

    \caption{Performance comparison of state-of-the-art text-to-video generation models on aesthetic score (Aes.\ Score), tag classification accuracy (Tag Cla.), and generation level (Gen.\ Level) metrics.}

    \label{tab:biao6}

\end{table}

\subsection{User Study}
We conducted a user study in the form of a questionnaire. Since most general users have not received professional training, and it is practically infeasible to train every participant in a large-scale survey, we only invited non-expert users to perform ranking evaluations on the outputs of 12 generative models across two dimensions: aesthetic quality and generation quality.

In the experiment, we randomly selected five prompt sets from the Prompt Suite for aesthetic quality and another five from the Prompt Suite for generation quality. For each prompt, we collected the corresponding videos generated by 12 different models, forming comparative sequences. Participants were asked to rank the videos according to two subjective yet representative criteria: ``To what extent does the video reflect the beauty described in the text?'' and ``How accurately does the video depict the textual content?''

As a reference, we normalize the evaluation scores across all sub-dimensions and compute their average to derive an overall ranking of the models. We then compare the overlap between human rankings and our model-based rankings using Recall@5, Recall@3, and Recall@1. The results from 40 collected questionnaires are summarized in Table~\ref{tab:biao7}.

\begin{table}

    \centering

    \small 

    \begin{tabular}{l c c c}

    \hline

     & Recall@1 & Recall@3 & Recall@5  \\

    \hline

    Aes.    & 0.10 & 0.70 & 0.80 \\

    Gen.    & 0.50 & 0.80 & 0.83 \\

    \hline

    \end{tabular}

    \caption{Comparison between human and model-based rankings on aesthetic and generation quality dimensions. Recall@5, Recall@3, and Recall@1 are reported based on 40 user surveys.}

    \label{tab:biao7}

\end{table}
\section{Conclusion}
We propose VGA-Bench, a fine-grained AIGC video evaluation benchmark comprising 52 sub-dimensions, 1,016 prompts, and over 60,000 annotated videos. Through our dedicated evaluators (VAQA-Net, VTag-Net, and VGQA-Net), this work delivers human-aligned insights into state-of-the-art models and systematically integrates artistic principles into the evaluation pipeline. Marking a paradigm shift from ``how real'' to ``how beautiful,'' VGA-Bench not only quantifies key elements like composition, color, and lighting, but also paves the way for models to achieve genuine perceptual aesthetics and expressiveness.
\maketitlesupplementary

\begin{strip}
    \centering
    \includegraphics[width=\textwidth]{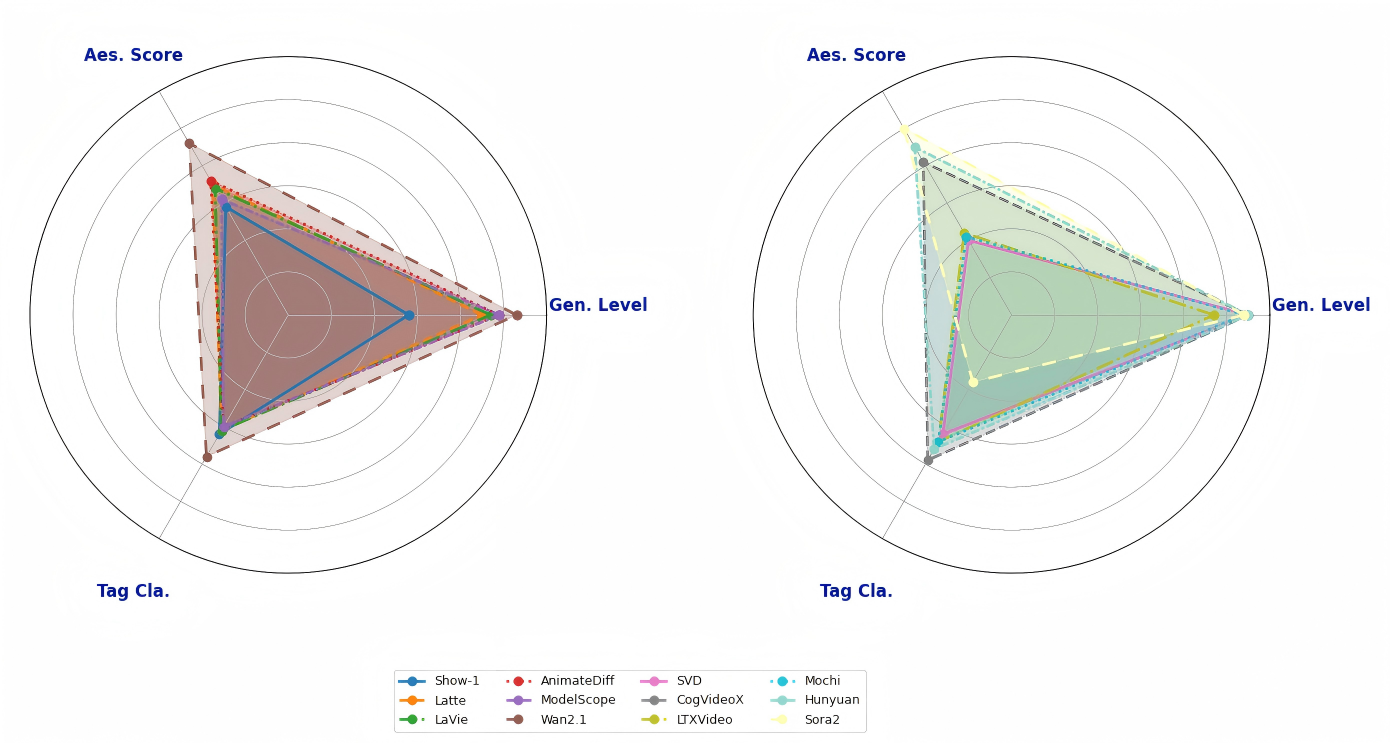}
    \captionof{figure}{Radar chart comparing the performance of various video generation models across three evaluation dimensions. The concentric circular grids represent score levels, ranging from 0.1 at the center to 0.9 outward in 0.1 intervals. Higher values indicate better performance. Each model is represented by a closed polygon with distinct colors and line styles for easy comparison.}
  \label{fig:radar}
\end{strip}

\section{Dimension of Generation Quality}
\label{sec:Generation}

The meanings of the dimensions of generation quality are summarized in Table \ref{tab:biao2}.

\section{Overview of Generative Model Performance}
\label{sec:radar}

The performance comparison of various generative models across the three core dimensions is shown in Figure \ref{fig:radar}.

\section{Comprehensive Evaluation Results of Generative Models across Sub-Dimensions in VGA-Bench}
\label{Sub-Dimensions}

\subsection{Performance Comparison of Generative Models on Aesthetic Quality Dimensions}
Table \ref{tab:biao8} presents the scoring results of 12 mainstream generative models across sub-attributes under the aesthetic quality dimension. All scores are generated by VAQA-Net through automated evaluation, reflecting the visual appeal and artistic expressiveness of the videos produced by these models.

\subsection{Comparison of Aesthetic Tag Prediction Capabilities}
Table \ref{tab:biao9} reports aesthetic tag prediction accuracies using VTag-Net, evaluating the models' capabilities in understanding and generating complex aesthetic semantics.

\subsection{Performance Comparison of Generative Models on the Generation Quality Dimension}
Due to the extensive sub-dimensions within generation quality, we conduct automated annotations using VGQA-Net and present the performance comparisons across three separate tables: Table \ref{tab:biao10} details 11 metrics on video-text consistency and spatio-temporal alignment; Table \ref{tab:biao11} assesses 14 realism metrics concerning physical laws and real-world commonsense; and Table \ref{tab:biao12} evaluates 6 technical dimensions reflecting basic low-level visual fidelity.

\begin{table*}[htbp]

    \centering

    \small

    \caption{Performance Comparison of Generative Models on Aesthetic Quality Dimensions}

    \label{tab:biao8}

    \begin{tabular}{ c c c c c c c c c c c c c}

        \hline

        \textbf{Model} & \textbf{Overall} & \textbf{Com} & \textbf{SS} & \textbf{Lig} & \textbf{VT} & \textbf{Col} & \textbf{DoF} & \textbf{Exp} & \textbf{Cos} & \textbf{Mak} \\  \hline
        Show-1~\cite{zhang2025show}     & 0.290 & 0.415 & 0.383 & 0.367 & 0.357 & 0.397 & 0.330 & 0.294 & 0.317 & 0.294 \\ 

        Latte-1~\cite{ma2024latte}      & 0.345 & 0.472 & 0.435 & 0.420 & 0.410 & 0.445 & 0.393 & 0.313 & 0.340 & 0.296 \\ 

        LaVie~\cite{wang2025lavie}      & 0.341 & 0.468 & 0.435 & 0.421 & 0.407 & 0.442 & 0.396 & 0.309 & 0.340 & 0.293 \\

        AnimateDiff~\cite{guo2023animatediff} & 0.356 & 0.475 & 0.446 & 0.427 & 0.415 & 0.460 & 0.400 & 0.317 & 0.364 & 0.310 \\

        ModelScope~\cite{wang2023modelscope}  & 0.312 & 0.445 & 0.413 & 0.397 & 0.381 & 0.430 & 0.377 & 0.287 & 0.334 & 0.266 \\

        Wan2.1~\cite{wan2025wan}        & \underline{0.459} & \underline{0.552} & \underline{0.523} & \underline{0.523} & \underline{0.521} & \underline{0.521} & \underline{0.503} & 0.436 & 0.459 & 0.412 \\ 

        SVD~\cite{blattmann2023stable}  & 0.204 & 0.305 & 0.260 & 0.229 & 0.233 & 0.277 & 0.191 & 0.207 & 0.151 & 0.164 \\ 

        CogVideoX~\cite{yang2024cogvideox} & 0.405 & 0.480 & 0.454 & 0.437 & 0.431 & 0.453 & 0.399 & 0.389 & 0.420 & 0.354 \\

        LTXVideo~\cite{hacohen2024ltx}  & 0.214 & 0.313 & 0.275 & 0.238 & 0.241 & 0.285 & 0.208 & 0.189 & 0.142 & 0.125 \\

        Mochi~\cite{genmo2024mochi}     & 0.211 & 0.306 & 0.269 & 0.240 & 0.234 & 0.283 & 0.203 & 0.198 & 0.147 & 0.147 \\

        Hunyuan~\cite{kong2024hunyuanvideo} & 0.452 & 0.531 & 0.505 & 0.507 & 0.504 & 0.512 & 0.485 & \underline{0.455} & \underline{0.465} & \underline{0.430} \\

        Sora2~\cite{liu2024sora}        & \textbf{0.504} & \textbf{0.559} & \textbf{0.532} & \textbf{0.543} & \textbf{0.564} & \textbf{0.560} & \textbf{0.540} & \textbf{0.520} & \textbf{0.532} & \textbf{0.452} \\
        \hline
    \end{tabular}

\end{table*}

\begin{table*}[htbp]
	\centering
    \small
	\caption{Comparison of Aesthetic Tag Prediction Capabilities}
	\label{tab:biao9}
	\begin{tabular}{*{12}{c}} 
        \hline 
        \textbf{Model} & \textbf{CT} & \textbf{NoLS} & \textbf{LSP} & \textbf{LQ} & \textbf{LC} & \textbf{ST} & \textbf{DoF} & \textbf{Sat} & \textbf{Bri} & \textbf{Col} & \textbf{Con} \\ \hline
        Show-1\cite{zhang2025show} & 0.17 & 0.47 & 0.19 & 0.46 & 0.17 & 0.32 & 0.54 & \underline{0.29} & 0.42 & 0.48 & 0.35 \\
        Latte-1\cite{ma2024latte} & 0.16 & 0.47 & 0.19 & 0.46 & 0.12 & 0.25 & 0.60 & \underline{0.29} & 0.35 & 0.46 & 0.35 \\
        LaVie\cite{wang2025lavie} & 0.17 & 0.47 & 0.19 & 0.46 & 0.11 & 0.25 & \underline{0.66} & \underline{0.29} & 0.41 & 0.39 & 0.35 \\
        AnimateDiff\cite{guo2023animatediff} & \underline{0.18} & 0.47 & 0.19 & 0.46 & 0.06 & 0.23 & \underline{0.66} & \underline{0.29} & 0.37 & 0.39 & 0.35 \\
        ModelScope\cite{wang2023modelscope} & 0.17 & 0.47 & 0.19 & 0.46 & 0.01 & 0.23 & 0.60 & \underline{0.29} & 0.43 & 0.40 & 0.35 \\
        Wan2.1\cite{wan2025wan} & \underline{0.18} & 0.60 & 0.19 & \textbf{0.57} & 0.23 & 0.34 & \textbf{0.68} & \textbf{0.38} & \underline{0.47} & 0.58 & \underline{0.36} \\
        SVD\cite{blattmann2023stable} & \textbf{0.19} & 0.50 & \underline{0.24} & 0.46 & 0.28 & 0.22 & 0.52 & \underline{0.29} & 0.25 & 0.46 & \textbf{0.37} \\
        CogVideoX\cite{yang2024cogvideox} & \underline{0.18} & \textbf{0.71} & 0.19 & \underline{0.51} & 0.33 & \underline{0.38} & 0.62 & \underline{0.29} & \textbf{0.48} & 0.63 & \textbf{0.37} \\
        LTXVideo\cite{hacohen2024ltx} & 0.15 & 0.52 & \textbf{0.27} & 0.46 & \underline{0.36} & 0.20 & 0.57 & \underline{0.29} & 0.25 & \underline{0.64} & \underline{0.36} \\
        Mochi\cite{genmo2024mochi} & 0.17 & 0.49 & \underline{0.24} & 0.46 & \textbf{0.44} & 0.20 & 0.55 & \underline{0.29} & 0.25 & \textbf{0.69} & 0.35 \\
        Hunyuan\cite{kong2024hunyuanvideo} & 0.16 & \underline{0.66} & 0.21 & 0.48 & 0.23 & \textbf{0.40} & \textbf{0.68} & 0.27 & 0.45 & 0.47 & 0.35 \\
        Sora2\cite{liu2024sora} & 0.13 & 0.32 & 0.14 & 0.29 & 0.19 & 0.32 & 0.27 & 0.11 & 0.15 & 0.19 & 0.10 \\ \hline
	\end{tabular}
\end{table*}

\begin{table*}[htbp]
	\centering
    \small
	\caption{Number and explanation of different assessment dimensions}
	\label{tab:biao2}
	\resizebox{1.9\columnwidth}{!}{
	\begin{tabular}{|c|c|c|p{8cm}|} 
		\hline 
		\textbf{Type} & \textbf{Num.} &\textbf{Assessment Dimension} & \textbf{Description} \\ \hline
		\multirow{11}{*}[-19ex]{\rotatebox[origin=c]{90}{Video-Text Consistency}} & \multirow{2}{*}{1} & \multirow{2}{*}{Character-Text Consistency} & Whether specific characters in the video match the text description (e.g., Elon Musk should appear as the correct individual). \\ \cline{2-4}
		& \multirow{3}{*}{2} & \multirow{3}{*}{Action-Text Consistency} & Whether actions in the video match the text description (e.g., running, jumping), focusing solely on the action regardless of the subject. \\ \cline{2-4}
		& \multirow{2}{*}{3} & \multirow{2}{*}{Scene-Text Consistency} & Whether scenes in the video match the described settings (e.g., hospital, school), including identifiable scene elements. \\ \cline{2-4}
		& \multirow{3}{*}{4} & \multirow{3}{*}{Object Position-Text Consistency} & Object positions refer to relative placement based on camera orientation (e.g., if ``a motorcycle is to the left of a bus,'' they should appear on corresponding sides of the video frame). \\ \cline{2-4}
		& \multirow{2}{*}{5} & \multirow{2}{*}{Object Attribute-Text Consistency} & Object attributes include descriptive features like color, shape, and texture. \\ \cline{2-4}
		& \multirow{2}{*}{6} & \multirow{2}{*}{Object-Text Consistency} & Whether objects in the video can be correctly identified as those mentioned in the text. \\ \cline{2-4}
		& \multirow{2}{*}{7} & \multirow{2}{*}{Video Content-Text Consistency} & Overall alignment where every textual description should be accurately generated. \\ \cline{2-4}
		& \multirow{2}{*}{8} & \multirow{2}{*}{Video Speed-Text Consistency} & Whether video speed matches textual descriptions (current samples only include slow-motion). \\ \cline{2-4}
		& \multirow{2}{*}{9} & \multirow{2}{*}{Video Style-Text Consistency} & Whether artistic styles mentioned in text (e.g., Van Gogh, Picasso) are recognizable in the video. \\ \cline{2-4}
		& \multirow{2}{*}{10} & \multirow{2}{*}{Camera Movement-Text Consistency} & Whether camera movements described in text (e.g., pan left, tilt right) are properly executed. \\ \cline{2-4}
		& \multirow{3}{*}{11} & \multirow{3}{*}{Unrealistic Description Imaginative Presentation} & When text describes unrealistic scenarios (e.g., ``an astronaut riding a horse in space''), whether the video presentation aligns with imaginative expectations. \\ \cline{2-4}
        
        \hline
		\multirow{14}{*}[-11ex]{\rotatebox[origin=c]{90}{Realism \& Plausibility}} & \multirow{2}{*}{12} & \multirow{2}{*}{Rigid Body Collision Realism} & Whether rigid body collisions in videos appear physically plausible. \\ \cline{2-4}
		& 13 & Action Realism & Whether actions could realistically be performed. \\ \cline{2-4}
		& \multirow{2}{*}{14} & \multirow{2}{*}{Scene Realism} & Whether scenes appear sufficiently realistic when no special style is specified in text. \\ \cline{2-4}
		& 15 & Weather Representation Realism & Whether weather conditions appear realistic. \\ \cline{2-4}
		& 16 & Time Period Representation Realism & Whether time-period representations appear authentic. \\ \cline{2-4}
		& \multirow{2}{*}{17} & \multirow{2}{*}{Gaseous Motion Realism} & Whether gas dynamics (smoke, vapor) appear physically accurate. \\ \cline{2-4}
		& 18 & Fluid Motion Realism & Whether fluid movements appear physically plausible. \\ \cline{2-4}
		& \multirow{2}{*}{19} & \multirow{2}{*}{Gradual Change Motion Realism} & Whether gradual transformations (balloon inflation, plant growth) appear physically accurate. \\ \cline{2-4}
		& \multirow{2}{*}{20} & \multirow{2}{*}{Object Motion Trajectory Realism} & Whether object movement paths follow physically plausible dynamics. \\ \cline{2-4}
		& 21 & Object Realism & Whether objects appear sufficiently realistic. \\ \cline{2-4}
		& 22 & Character Generation Quality & Whether human characters appear sufficiently realistic. \\ \cline{2-4}
		& \multirow{2}{*}{23} & \multirow{2}{*}{Textual Attribute Representation Realism} & Whether object attributes (color, shape, texture) match real-world appearances. \\ \cline{2-4}
		& 24 & Video Lighting and SGQAow Realism & Whether lighting and sGQAows appear physically accurate. \\ \cline{2-4}
		& \multirow{2}{*}{25} & \multirow{2}{*}{Moving Scene Reasonableness} & Whether scene transitions during camera movements maintain proper perspective. \\
        \cline{2-4}
        & 26 & Overall Realism & Whether the entire video looks realistic overall. \\
        \hline
		\multirow{7}{*}[-1ex]{\rotatebox[origin=c]{90}{Basic Quality}} 
        & \multirow{2}{*}{27} & \multirow{2}{*}{Abnormal Lighting Detection} & Videos should avoid lighting artifacts (overexposure, abnormal flares). \\ \cline{2-4}
		& 28 & Video Noise-Free & Videos should exhibit no noticeable noise artifacts. \\ \cline{2-4}
		& 29 & Video Clarity & Whether video resolution is sufficiently sharp. \\ \cline{2-4}
		& \multirow{2}{*}{30} & \multirow{2}{*}{Static Content Non-distortion} & Stationary objects shouldn't distort abnormally during camera movement. \\ \cline{2-4}
		& \multirow{2}{*}{31} & \multirow{2}{*}{Static Content Stability} & Stationary objects shouldn't distort abnormally over time (temporal consistency). \\ \cline{2-4}
		\hline
	\end{tabular}}
\end{table*}

\begin{table*}[htbp]
	\centering
    \small
	\caption{Evaluation Results on Video-Text Consistency}
	\label{tab:biao10}
	\begin{tabular}{*{12}{c}} 
        \hline 
        \textbf{Model} & \textbf{1} & \textbf{2} & \textbf{3} & \textbf{4} & \textbf{5} & \textbf{6} & \textbf{7} & \textbf{8} & \textbf{9} & \textbf{10} & \textbf{11} \\ \hline
        Show-1\cite{zhang2025show} & 0.037 & 0.364 & 0.362 & 0.346 &0.505 & 0.466 & 0.196 & 0.027 & 0.030 & 0.433 & 0.250  \\
        Latte-1\cite{ma2024latte} & 0.052 & 0.285 & 0.770 & 0.565 & 0.875 & 0.672 & 0.598 & 0.000 & 0.034 & 0.500 & 0.288  \\
        LaVie\cite{wang2025lavie} & \textbf{0.076} & 0.212 & 0.852 & 0.512 & 0.887 & 0.682 & 0.598 & 0.007 & 0.064 & 0.612 & 0.340  \\
        AnimateDiff\cite{guo2023animatediff} & 0.024 & 0.168 & 0.783 & 0.500 & \textbf{0.977} & 0.770 & 0.547 & 0.000 & 0.014 & \textbf{0.861} & 0.400  \\
        ModelScope\cite{wang2023modelscope} & 0.021 & 0.208 & 0.754 & 0.464 & 0.964 & 0.816 & 0.567 & 0.014 & 0.037 & \underline{0.806} & \textbf{0.480}  \\
        Wan2.1\cite{wan2025wan} & 0.026 & 0.840 & 0.672 & \underline{0.629} & 0.900 & 0.795 & \underline{0.628} & \textbf{0.295} & \textbf{0.184} & 0.595 & \underline{0.448}  \\
        SVD\cite{blattmann2023stable} & \underline{0.066} & \textbf{0.958} & \textbf{0.893} & 0.572 & 0.934 & \textbf{0.917} & 0.591 & 0.071 & 0.050 & 0.655 & 0.410  \\
        CogVideoX\cite{yang2024cogvideox} & 0.055 & \underline{0.936} & \underline{0.863} & 0.512 & 0.929 & \underline{0.888} & 0.599 & 0.045 & 0.065 & 0.720 & 0.419  \\
        LTXVideo\cite{hacohen2024ltx} & 0.032 & 0.561 & 0.578 & 0.542 & 0.781 & 0.640 & 0.538 & 0.038 & 0.043 & 0.533 & 0.303  \\
        Mochi\cite{genmo2024mochi} & 0.039 & 0.897 & 0.803 & \textbf{0.701} & 0.951 & 0.813 & \textbf{0.679} & 0.039 & 0.043 & 0.632 & 0.387  \\
        Hunyuan\cite{kong2024hunyuanvideo} & 0.056 & 0.856 & 0.799 & 0.592 & \underline{0.970} & 0.809 & 0.600 & 0.034 & 0.041 & 0.620 & 0.350  \\
        Sora2\cite{liu2024sora} & 0.024 & 0.842 & 0.790 & 0.532 & 0.934 & 0.851 & 0.603 & \underline{0.086} & \underline{0.094} & 0.715 & 0.425 \\ \hline
	\end{tabular}
\end{table*}

\begin{table*}[htbp]
	\centering
    \footnotesize
	\caption{Evaluation Results on Realism \& Plausibility}
	\label{tab:biao11}
	\begin{tabular}{*{16}{c}}
        \hline
        \textbf{Model} & \textbf{12} & \textbf{13} & \textbf{14} & \textbf{15} & \textbf{16} & \textbf{17} & \textbf{18} & \textbf{19} & \textbf{20} & \textbf{21} & \textbf{22} & \textbf{23} & \textbf{24} & \textbf{25} & \textbf{26} \\ \hline
        Show-1\cite{zhang2025show} &0.028 &0.267 &0.294 &0.377 &0.206 &0.344 &0.377 &0.609 &0.127 &0.351 &0.237 &0.389 &0.284 &0.282 & 0.227 \\
        Latte-1\cite{ma2024latte} &0.038 &0.435 &0.511 &0.640 &0.388 &0.350 &0.488 &0.825 &0.150 &0.428 &0.513 &0.600 &0.413 &0.463 &0.598 \\
        LaVie\cite{wang2025lavie} &0.035 &0.402 &0.529 &0.618 &0.345 &0.415 &0.490 &0.830 &0.145 &0.408 &0.530 &0.607 &0.440 &0.533 &0.598 \\
        AnimateDiff\cite{guo2023animatediff} &0.025 &0.490 &0.535 &0.644 &0.415 &0.390 &0.500 &0.760 &0.145 &0.466 &0.611 &0.677 &0.485 &0.556 & 0.547 \\
        ModelScope\cite{wang2023modelscope} &0.070 &0.484 &0.522 &0.676 &0.405 &0.375 &0.490 &0.870 &0.180 &0.508 &0.574 &0.660 &\underline{0.490} &0.574 &0.567 \\
        Wan2.1\cite{wan2025wan} &\textbf{0.163} &0.517 &0.521 &0.539 &\textbf{0.481} &0.575 &0.500 &\textbf{0.938} &\textbf{0.210} &0.527 &0.521 &0.572 &\textbf{0.494} &0.533 &0.628 \\
        SVD\cite{blattmann2023stable} &0.025 &0.496 &\textbf{0.586} &0.672 &0.445 &\textbf{0.655} &0.500 &0.660 &0.165 &\underline{0.605} &0.652 &0.640 &0.440 &\textbf{0.671} & 0.591 \\
        CogVideoX\cite{yang2024cogvideox} &0.030 &0.512 &0.551 &\underline{0.677} &0.461 &\underline{0.597} &0.470 &0.875 &0.139 &0.590 &0.631 &0.602 &0.420 &0.641 & 0.599 \\
        LTXVideo\cite{hacohen2024ltx} &0.035 &0.465 &0.520 &0.605 &0.391 &0.471 &0.444 &0.864 &0.131 &0.497 &0.520 &0.595 &0.424 &0.509 & 0.623 \\
        Mochi\cite{genmo2024mochi} &0.047 &0.506 &0.540 &0.665 &0.415 &0.500 &\textbf{0.560} &\underline{0.937} &0.174 &0.535 &\underline{0.653} &0.631 &0.466 &0.578 & \textbf{0.786} \\
        Hunyuan\cite{kong2024hunyuanvideo} &0.035 &0.526 &\underline{0.563} &\textbf{0.784} &0.428 &\textbf{0.526} &\underline{0.547} &0.864 &0.172 &0.593 &\textbf{0.714} &\textbf{0.700} &0.438 &0.616 & \underline{0.694}\\
        
        Sora2\cite{liu2024sora} &\underline{0.085} &\underline{0.520} &0.553 &0.660 &\underline{0.450} &0.585 &0.490 &0.780 &\underline{0.190} &\textbf{0.620} &0.604 &\underline{0.694} &0.450 &\underline{0.668} & 0.603 \\ \hline
	\end{tabular}
\end{table*}

\begin{table*}[htbp]
	\centering
	\caption{Evaluation Results on Basic Visual Quality}
	\label{tab:biao12}
	\begin{tabular}{*{6}{c}} 
        \hline 
        \textbf{Model} & \textbf{27} & \textbf{28} & \textbf{29} & \textbf{30} & \textbf{31}  \\ \hline
        Show-1\cite{zhang2025show} &0.369 &0.097 &0.212 &0.257 &0.285 \\
        
        Latte-1\cite{ma2024latte} &0.766 &0.158 &0.365 &0.447 &0.742  \\
        
        LaVie\cite{wang2025lavie} &0.861 &0.207 &0.452 &0.500 &0.717\\
        AnimateDiff\cite{guo2023animatediff} &0.843 &0.214 &0.526 &\underline{0.525} &0.680  \\
        ModelScope\cite{wang2023modelscope} &0.633 &0.232 &0.454 &0.482 &0.664 \\
        
        SVD\cite{blattmann2023stable} &\underline{0.898} &\underline{0.292} &\textbf{0.650} &0.511 &0.733 \\
        CogVideoX\cite{yang2024cogvideox} &0.855 &\textbf{0.304} &\underline{0.592} &0.515 &0.748 \\
        LTXVideo\cite{hacohen2024ltx} &0.823 &0.238 &0.476 &0.518 &\textbf{0.808} \\
        
        Mochi\cite{genmo2024mochi} &0.824 &0.223 &0.500 &0.511 &\underline{0.805} \\
        Hunyuan\cite{kong2024hunyuanvideo} &0.876 &0.253 &0.556 &\textbf{0.578} &0.797 \\
        
        Sora2\cite{liu2024sora} &\textbf{0.906} &0.215 &0.540 &0.518 &0.720 \\ \hline
	\end{tabular}
\end{table*}
\clearpage
\newpage

\small
\section*{Acknowledgments}
This work was supported by the Ant Group Research Fund, the National Natural Science Foundation of China under Grant No. 62072014, and the Opening Project of the State Key Laboratory of General Artificial Intelligence, BIGAI/Peking University, Beijing, China (Project No. SKLAGI2025OP01).

\newpage
\bibliographystyle{ieeenat_fullname}
\bibliography{main}


\end{document}